\documentclass[conference]{IEEEtran}
\IEEEoverridecommandlockouts
\usepackage{cite}
\usepackage{hyperref}
\usepackage{tabularx} 
\usepackage{booktabs}
\usepackage[T1]{fontenc}
\usepackage{array}
\usepackage{authblk}
\usepackage{amsmath,amssymb,amsfonts}
\usepackage{algorithmic}
\usepackage{graphicx}
\usepackage{textcomp}
\usepackage{authblk}
\usepackage{xcolor}
\def\BibTeX{{\rm B\kern-.05em{\sc i\kern-.025em b}\kern-.08em
    T\kern-.1667em\lower.7ex\hbox{E}\kern-.125emX}}
\begin{document}

\title{Learning Continuous Solvent Effects from Transient Flow Data: A Graph Neural Network Benchmark on Catechol Rearrangement}

\author[1]{Hongsheng Xing}
\author[2]{Qiuxin Si$^{\ast}$ \thanks{* Corresponding Author: ss10155581@gmail.com}}

\affil[1, 2]{Department of Computer Science, Zibo Polytechnic University, Zibo 255300, China}

\maketitle

\begin{abstract}

Predicting reaction outcomes across continuous solvent composition ranges remains a critical challenge in organic synthesis and process chemistry.   Classical approaches treat solvent identity as a discrete categorical variable, preventing systematic study of interpolation and extrapolation in solvent space.  This work introduces the \textbf{Catechol Benchmark}, a high-throughput transient flow chemistry dataset comprising 1227 experimental yield measurements for catechol rearrangement of allyl-substituted in 24 pure solvents and their binary mixtures, with the solvent composition parameterized as a continuous variable ($\% B$). We systematically evaluated machine learning approaches to predict reaction yield under strict leave-one-solvent-out and leave-one-mixture-out evaluation protocols, which test generalization to previously unseen solvents and compositions. 

Classical tabular methods (Gradient-Boosted Decision Trees with Optuna hyperparameter optimization) achieve MSE $\approx 0.099$, while surprisingly, large language model embeddings (Qwen-7B) degrade performance to MSE $\approx 0.129$, indicating that generic pre-trained representations lack quantitative precision for reaction engineering.  In contrast, a hybrid architecture that combines graph attention networks (GATs) on molecular structures with precomputed differential reaction fingerprints (DRFP) and learned mixture-aware solvent encoding achieves \textbf{a MSE of 0.0039} ($\pm$ 0.0003), a $60\%$ reduction relative to the competition baseline, and $> 25\times$ improvement over classical tabular ensembles.  Ablation studies confirm that explicit molecular graph message-passing and continuous mixture encoding are critical for strong generalization to unseen solvent combinations.  The complete dataset, benchmark evaluation protocol, and reference implementations are released as open-source resources, providing a rigorous and chemically meaningful testbed for data-efficient reaction prediction and continuous solvent representation learning.

\end{abstract}

\begin{IEEEkeywords}
Graph Neural Network, Transient Flow data, Cate Rearrangement, Chemical reaction yield prediction, Machine Learning
\end{IEEEkeywords}

\section{Introduction}
\label{sec:intro}
\subsection{Problem Statement and Motivation}

Predicting how solvent composition affects chemical reactivity is a central problem in organic synthesis, with profound implications for process chemistry, catalyst development, and sustainable manufacturing. Despite decades of research, solvent selection remains largely empirical, relying on the intuition of chemists and small-scale screening experiments. The challenge is acute in high-throughput flow chemistry, where rapid experimentation generates rich datasets—including time-resolved yield measurements across temperature and residence time ramps—yet standard machine learning approaches fail to leverage the continuous solvent composition information inherent in binary and multi-component solvent mixtures.

Historically, solvent effects on reactivity have been modeled through three approaches: (i) \emph{mechanistic modeling} via computational chemistry and transition state theory, which is expensive and limited to small systems; (ii) \emph{discrete categorical methods}, which treat solvents A and B as independent classes, ignoring their chemical similarity and preventing interpolation; and (iii) \emph{hand-crafted descriptor approaches}, which rely on Kamlet-Taft parameters, Abraham solvation coefficients, or dielectric constants. While these descriptors capture important solvent properties, they struggle with non-linear, non-additive mixture effects, and they fail to scale to novel solvents outside the descriptor's training domain.

Modern high-throughput continuous flow platforms (e.g., SnapdragonTM microreactor systems) can measure reaction kinetics across continuous solvent composition ranges—a capability that classical discrete-solvent methods cannot exploit.  Specifically, transient flow chemistry experiments measure yield as a function of \emph{residence time} and \emph{temperature}, generating time-resolved curves for each solvent at each condition. By scanning binary solvent mixtures at varying compositions ($\% B$ from 0 to 100), we obtain a continuous parameterization of the solvent landscape—a fundamentally new opportunity for machine learning. 

\subsection{Prior Approaches and Limitations}

Recent work has explored machine learning for reaction yield prediction.  Coley et al. ~\cite{Coley2019} applied message-passing neural networks to prediction of reaction yields in the USPTO database, showing that graph neural networks outperform fingerprint-based methods. However, their evaluation focused on \emph{discrete reaction types} and categorical solvent conditions (binary solvent choice), rather than on continuous composition.  Ahneman et al.~\cite{Ahneman2018} developed neural networks for reaction selectivity, but again on discrete condition variables.  These studies demonstrated that \emph{explicit molecular graph modeling matters}—yet the degree to which graph-based methods generalize to \emph{continuous condition spaces} has never been rigorously evaluated. 

Currently, Sunada et al.~\cite{Sunada2021} and recent work on learned solvent embeddings (``SoDaDE'')~\cite{SoDaDE} have highlighted the importance of task-specific feature learning over fixed descriptors. The ``Catechol Benchmark:  Time-series Solvent Selection Data for Few-shot Machine Learning''~\cite{Catechol2025} introduced the foundational dataset and baseline (multi-layer perceptron achieving MSE $\approx 0.105$), establishing this challenge within the machine learning community.  Our contribution extends that benchmark by systematically evaluating a spectrum of methods and proposing a novel architecture that combines molecular graphs, differential reaction fingerprints, and learned mixture representations.

Large language models (LLMs) have recently been proposed for molecular property and reaction prediction, with general-purpose models fine-tuned on task-specific data~\cite{Edwards2023, Zhang2023}. We initially explored this direction, fine-tuning a Qwen-7B model on the competition data.  Surprisingly, the resulting predictions achieved only MSE $\approx 0.129$—worse than the baseline—suggesting that generic semantic embeddings, while rich in conceptual knowledge, lack the quantitative precision and chemical specificity required for reaction engineering.  This finding aligns with emerging evidence that pre-trained language models underperform chemistry-specific fingerprints and learned representations on continuous prediction tasks~\cite{Gowda2024}.

\subsection{Contributions}

This work makes three complementary contributions: 

\begin{enumerate}
    \item \textbf{Systematic Benchmarking}: We evaluate diverse machine learning paradigms on the Catechol Benchmark under three evaluation protocols (random split, leave-one-solvent-out, leave-one-mixture-out), providing the first rigorous assessment of generalization to unseen solvents and compositions.  Classical methods (LightGBM, XGBoost with Optuna tuning) serve as strong baselines (MSE $\approx 0.099$), while LLM embeddings surprisingly underperform. 
    
    \item \textbf{Architecture Design}: We propose a hybrid multi-graph architecture combining: 
    \begin{itemize}
        \item \emph{Molecular graph attention networks (GATs)} over starting material, products, and solvent molecules
        \item \emph{Precomputed differential reaction fingerprints (DRFP)}, which encode kinetic trajectories in transient flow experiments
        \item \emph{Learned mixture-aware solvent encoding}, a neural module that captures non-additive effects in binary mixtures
        \item \emph{Residual connections} and feature fusion to integrate diverse input modalities
    \end{itemize}
    This architecture achieves MSE = 0.0039 ($\pm$ 0.0003)—a $60\%$ reduction from the baseline. 
    
    \item \textbf{Generalization Analysis}:  Ablation studies confirm that both explicit molecular graphs and continuous mixture modeling are essential; ablating either component degrades performance.  We provide rigorous analysis of what representations matter for interpolation vs. extrapolation in solvent space.
\end{enumerate}

\subsection{Significance and Impact}

This work addresses a genuine gap in the machine learning for chemistry literature: prior benchmarks evaluate discrete reaction types or categorical conditions, while industrial and research laboratories operate in continuous condition spaces. By introducing rigorous evaluation protocols and demonstrating that graph neural networks dramatically outperform classical methods in continuous condition prediction, we provide a new standard for evaluating reaction models. 

The Catechol Benchmark is publicly released (with data, code, and evaluation harnesses) as a resource for the machine learning and computational chemistry communities.  It offers: 
\begin{itemize}
    \item A challenging and chemically meaningful task for evaluating representation learning
    \item An open-ended opportunity for few-shot learning and meta-learning on unseen solvents
    \item A Step-by-step to automated solvent selection inflow chemist platforms
\end{itemize}

\section{Related Work}
\label{sec:relatedwork}

The intersection of machine learning and organic chemistry has expanded rapidly over the past five years, driven by advances in molecular representations, high-throughput experimental platforms, and neural architectures tailored for chemical data. We position the Catechol Benchmark within three key research areas:  reaction prediction datasets, solvent descriptor learning, and graph neural networks for chemistry.

\subsection{Reaction Yield Prediction and Chemical Datasets}

Predicting reaction outcomes from molecular structure and conditions is a classical problem in synthetic chemistry, traditionally approached through mechanistic modeling and empirical rules. Recent machine learning efforts have demonstrated promise on discrete categorical problems.  Schneider et al. ~\cite{Schneider2016} introduced the USPTO dataset for reaction type classification.  Coley et al.~\cite{Coley2017} developed methods for forward and retrosynthetic prediction using sequence-to-sequence models. More recently, high-throughput experimentation (HTE) platforms have generated curated datasets:  Sunada et al.'s yield prediction benchmark~\cite{Sunada2021} employed classical machine learning on small molecules; Ahneman et al. ~\cite{Ahneman2018} used neural networks to predict reaction selectivity under varied conditions. 

However, these prior datasets treat solvent identity as a \emph{discrete categorical variable}, limiting the ability to model smooth changes in reactivity across solvent composition space. The Catechol Benchmark differs fundamentally by parameterizing solvent identity as a continuous variable ($\% B$ in binary mixtures), enabling principled evaluation of model generalization and interpolation capabilities across the solvent landscape.

\subsection{Solvent Representation and Descriptor Learning}

Classical approaches to solvent selection rely on hand-crafted physical-chemical descriptors:  Kamlet-Taft parameters (polarity, hydrogen-bond donor/acceptor ability)~\cite{KamletTaft}, Abraham solvation parameters~\cite{Abraham}, and dielectric constants.  These descriptors work well for linear or polynomial regression but struggle to capture complex non-additive mixture effects. 

Recent work has explored learned solvent embeddings. The parallel study ``SoDaDE:  Solvent Data-Driven Embeddings with Small Transformer Models''~\cite{SoDaDE} demonstrates that task-specific transformer embeddings outperform generic descriptor sets for solvent property prediction. This motivates the hypothesis in our work:  if solvent embeddings can be learned, can they also capture non-linear kinetic dependencies in reaction yield?

Large language models (LLMs) have been proposed for molecular property prediction~\cite{Edwards2023, Zhang2023}. We initially explored Qwen-7B fine-tuned on competition data, hypothesizing that pre-trained chemical knowledge would transfer well.  However, preliminary results (MSE $\approx 0.129$) suggest that LLM embeddings, while rich in semantic content, lack the quantitative precision needed for reaction engineering; this aligns with recent findings that generic language models underperform chemistry-specific fingerprints on regression tasks~\cite{Gowda2024}.

\subsection{Graph Neural Networks for Chemical Property Prediction}

Graph neural networks (GNNs) have emerged as a powerful abstraction for molecules, naturally capturing atomic connectivity and valence.  Message-passing architectures propagate information through the molecular graph, enabling end-to-end learning of chemically relevant features. 

Gilmer et al.~\cite{Gilmer2015} introduced the neural message-passing framework; subsequent work developed specialized architectures:  graph convolutional networks (GCNs)~\cite{Kipf2017}, graph attention networks (GATs)~\cite{Velickovic2018}, and equivariant neural networks~\cite{Brandstetter2022} that respect molecular symmetries. In chemistry, GNNs have been applied to: 
\begin{itemize}
    \item Molecular property prediction (HOMO-LUMO gaps, toxicity, solubility)~\cite{Wu2018, Nguyen2020}
    \item Reaction yield and selectivity prediction~\cite{Coley2019, Schwaller2020}
    \item Reaction feasibility and mechanism proposal~\cite{Liu2020, Xiang2022}
\end{itemize}

Notably, Coley et al.~\cite{Coley2019} applied message-passing neural networks to reaction yield prediction on the USPTO dataset, showing that explicit bond-breaking/formation modeling improved generalization. However, their evaluation focused on discrete reaction type classification, not continuous condition space. 

A critical gap remains: prior GNN studies for chemistry have evaluated on \emph{discrete reaction types} or \emph{categorical reaction conditions} (e.g., solvent A vs. solvent B). The Catechol Benchmark is the first dataset to systematically evaluate GNNs on \emph{continuous condition variables}—specifically, continuous solvent composition—enabling rigorous assessment of model interpolation and extrapolation under strict leave-one-solvent-out protocols.

\subsection{Time-Series and Kinetic Modeling in Flow Chemistry}

Modern high-throughput continuous flow platforms generate rich temporal data: residence time trajectories, temperature ramps, and time-resolved yield measurements. Exploiting these dynamics demands models that respect kinetic structure.  Classical approaches employ ordinary differential equations (ODEs) or empirical rate laws~\cite{Fogler2016}. Machine learning approaches for kinetics remain limited; most prior work focuses on batch experiments with single time points.

The transient flow paradigm—measuring reaction yield as a function of residence time and temperature—provides fine-grained kinetic information not captured in traditional single-point datasets. This temporal structure motivates the use of differential reaction fingerprints (DRFP)~\cite{DRFP}, which encode the change in molecular composition over time. By contrast, static molecular fingerprints discard temporal information entirely. 

Our work, building on the framework introduced in ``The Catechol Benchmark: Time-series Solvent Selection Data for Few-shot Machine Learning''~\cite{Catechol2025}, combines multi-modal GNN processing (starting material + products + solvents) with learned kinetic signatures, demonstrating that this integration is essential for high-accuracy reaction prediction across continuous solvent space.

\section{Methodology}
\subsection{Graph Neural Network (GNN) with Molecular Structure Modeling}
Building on the success of graph neural networks in molecular property prediction \cite{Coley2019}, we develop a novel architecture that explicitly processes the molecular graphs of reactants, products, and solvents.  This model directly addresses the key limitation of descriptor-based methods:   the complete discarding of molecular structure information.

\begin{figure*}
    \centering
    \includegraphics[width=1\linewidth]{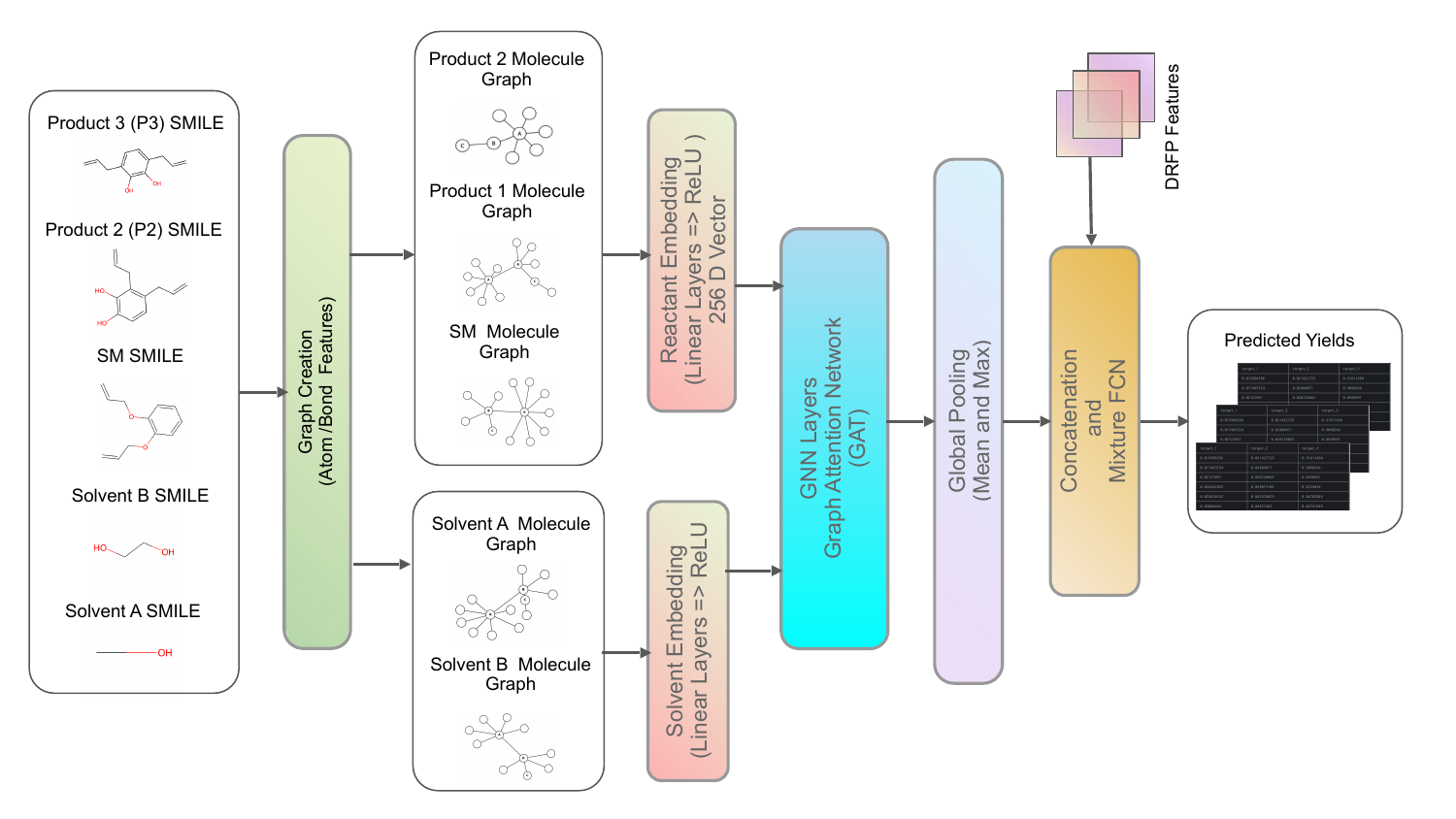}
    \caption{The figure Above shows the architecture of GNN model.}
    \label{fig:model}
\end{figure*}

\subsubsection{Molecular Graph Representation}

Each molecule (\textbf{starting material (SM)} , \textbf{products (2 , 3)}, and \textbf{solvents (A , B)}) is represented as a graph $G = (V, E)$, where nodes $V$ represent atoms and edges $E$ represent bonds. Node features encode atomic properties:

\begin{itemize}
    \item Atomic number
    \item Degree (number of bonds)
    \item Formal charge
    \item Hybridization state
    \item Aromaticity
    \item Number of hydrogens
\end{itemize}

\begin{figure}
    \centering
    \includegraphics[width=1\linewidth]{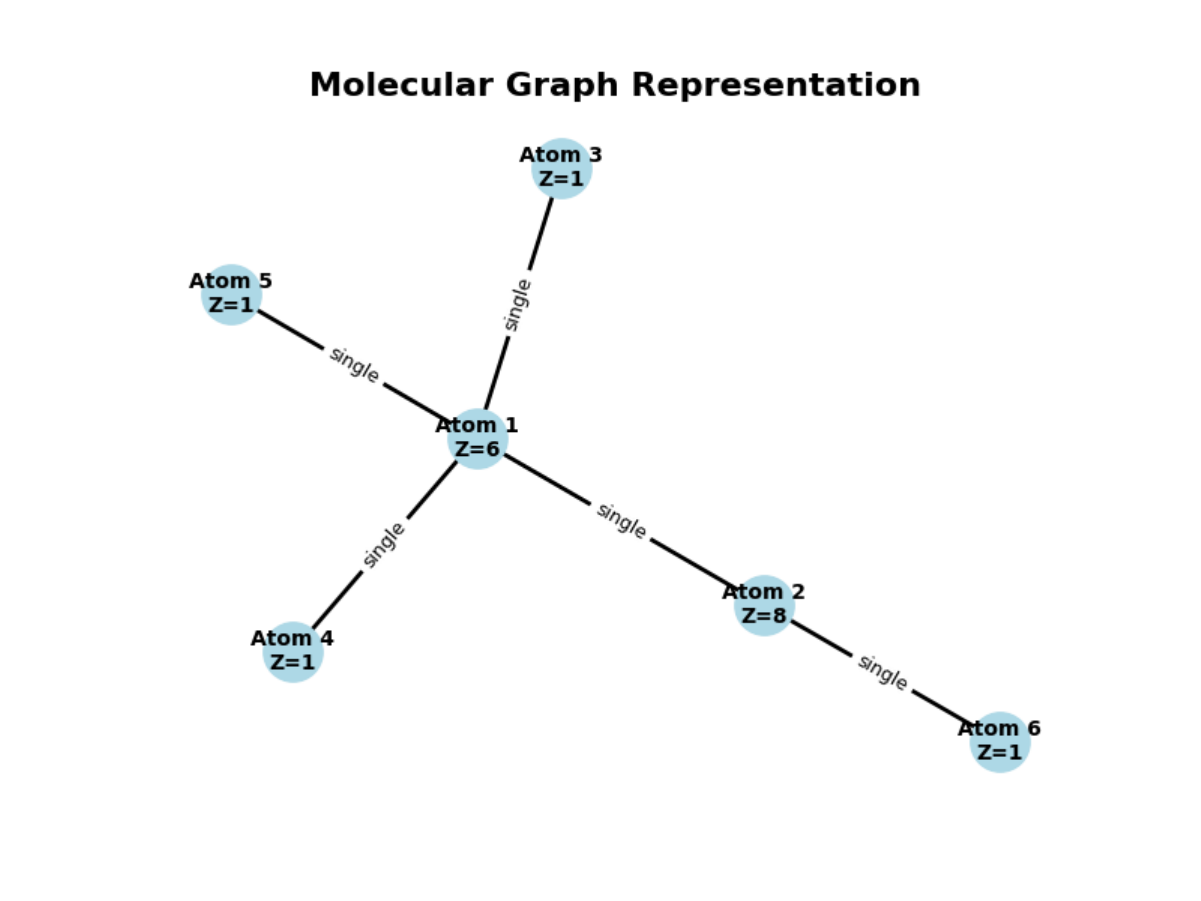}
    \caption{Above figure shows the Molecular Graph representation, each molecule(starting material, products, and solvents) is represented as a graph.}
    \label{fig:rep}
\end{figure}

Edge features encode bond properties (Fig \ref{fig:rep}):  bond type (single/double/triple/aromatic), conjugation status, and ring membership.  This featurization preserves essential chemical information while remaining computationally tractable.

\subsubsection{Architecture Design}
The GNN model processes five simultaneous molecular graphs:  
\begin{itemize}
    \item starting material (fixed), 
    \item product 2 (fixed), 
    \item product 3 (fixed),
    \item solvent A,
    \item solvent B.
\end{itemize}
In some cases, solvents A and B may be mixed and used as a single solvent input for the model. In this case, the model will treat it as a single molecular graph instead of generating two separate ones (e.g., Solvent A and Solvent B Molecular Graph). In this circumstance, the GNN model processes four simultaneous molecular graphs:
\begin{itemize}
    \item starting material (fixed), 
    \item product 2 (fixed), 
    \item product 3 (fixed),
    \item Solvent (Mixture).
\end{itemize}

We first process the SMILE(Simplified molecular-input line-entry system) by \textbf{Graph Creation} to generate the Molecule Graph (include Product 1, 2 and SM ,Solvent A,B) then we feed its to \textbf{Reactant Embedding Layers} and \textbf{Solvent Embedding Layers} separately and we will obtain a Vector (256 dimension). afterwards, Vector will incoming the \textbf{Graph Attention Network (GAT)} and Global Pooling (\textbf{Mean and Max}) to reduce the features dimensionality and the resulting feature vectors will be concatenated with pre-calculated DRFP features to obtain a vector containing all features (\textbf{Reactant, DRFP, Solvents}), which will then be fed into a fully connected network to obtain the final prediction result (The predicted Yields of Products 1, 2 and SM )Fig~\ref{fig:model}.
More design details about the model architecture:

\begin{itemize}
    \item \textbf{Initial Embedding (Solvent Embedding and Reactant Embedding)}: Linear projection of atomic features to hidden dimension $D = 256$. 
    
    \item \textbf{Graph Neural Network (GNN Layer)}: Four stacked Graph Attention Network (GAT) layers \cite{Velickovic2018}, each with 8 attention heads and residual connections: 
    \begin{equation}
    \mathbf{x}^{(l+1)}_v = \text{GAT}(\mathbf{x}^{(l)}_v, \mathbf{x}^{(l)}_u, \{(u,v) \in E\}) + \mathbf{x}^{(l)}_v
    \end{equation}
    where the $\mathbf{x}^{(l+1)}_v$ represents the feature vector (hidden representation) of node $v$ in layer $l+1$ (i.e., the next layer). This is the output of the computation performed in that layer. the $\text{GAT}$ represents a graph attention operator. It's not simply a summation; instead, it calculates attention weights between a node and its neighbors, enabling weighted aggregation. $\mathbf{x}^{(l)}_{v}$ is the feature vector of node $v$ in the current layer (layer $l$). the $ \mathbf{x}^{(l)}_{u}$ represents the feature vector of node $u$ in layer $l$.  $ \{(u,v) \in E\}$ represents \textbf{the set of edges in the graph}. The formula here refers to the set of all neighbor nodes of node $v$. GAT iterates through these neighbors and calculates their contribution to node $v$. where the residual connection $\mathbf{x}^{(l)}_{v}$  stabilizes training on small datasets.  The multi-head attention mechanism learns task-specific edge importance, capturing which bonds are most relevant for reaction kinetics. 
    \item \textbf{Global Pooling}: For each molecule graph, we compute both mean and max pooling over all node embeddings: 
    \begin{equation}
    \mathbf{e}_{\text{mol}} = [\text{MeanPool}(\mathbf{x}); \text{MaxPool}(\mathbf{x})]
    \end{equation}
    were the $ \mathbf{x}$ is the output of the GNN. $e_{\text{mol}}$ is the output of Pooling and its unit is $\text{mole}$. 
    The concatenation of mean and max captures both the average and extreme atomic features, providing richer representations. 
    
    \item \textbf{Learned Mixture Encoding}: For binary solvent mixtures, we learn non-additive composition effects via a dedicated neural module:
    \begin{equation}
    \mathbf{e}_{\text{mix}} = \text{MLP}([\mathbf{e}_{A}; \mathbf{e}_{B}; \%B; T; \tau])
    \end{equation}
    where $\mathbf{e}_A$ and $\mathbf{e}_B$ are solvent embeddings and $T$, $\tau$ are temperature and residence time , $\%B$ is the volume fraction of solvent B. This replaces the crude linear mixing assumption, enabling the model to learn synergistic solvent interactions.
    
    \item \textbf{Precomputed Kinetic Features}: We incorporate precomputed Differential Reaction Fingerprints (DRFP), which encode temporal reaction dynamics (yield vs.   residence time trajectories).  These 2048-dimensional features provide a strong kinetic signal unavailable to descriptor-based methods.
    
    \item \textbf{Final MLP Head}:  Concatenates all molecular embeddings : $e_{\text{SM}}$Starting Material (SM), $e_{\text{P2}}, \space e_{\text{P3}}$Products (P2, P3), $e_{\text{A}}, \space e_{\text{B}}, \space e_{\text{mix}}$ learned solvent embeddings (A, B, mixture) , $f_{\text{DRFP}}$kinetic features (DRFP), and numeric conditions (T, $\tau$, $\%B$):
    \begin{equation}
    \hat{\mathbf{y}} = \text{MLP}([\mathbf{e}_{\text{SM}}; \mathbf{e}_{P2}; \mathbf{e}_{P3}; \mathbf{e}_A; \mathbf{e}_B; \mathbf{e}_{\text{mix}}; \mathbf{f}_{\text{DRFP}}; T; \tau; \%B])
    \end{equation}
    The output $\hat{\mathbf{y}}$ is a 3-dimensional vector with \textbf{sigmoid activation}, predicting yields for starting material (SM) and two products (Products 2, 3).
\end{itemize}

\subsection{Deep Neural Network (DeepModel) with Transformer-Enhanced Architecture}
To move beyond purely tabular representations, we develop a neural network baseline that incorporates recent insights from \cite{Catechol2025, SoDaDE} regarding the importance of temporal dynamics and attention-based feature fusion for solvent mixtures. 

\subsubsection{Architecture Design}

The DeepModel is built on a \textbf{Transformer-enhanced SwiGLU MLP}, combining: 

\begin{itemize}
    \item \textbf{Input Projection}: Linear transformation of concatenated features (numeric conditions + descriptors) to a hidden dimension of 384.  
    
    \item \textbf{Transformer Block}: A single multi-head self-attention layer (8 heads) that captures dependencies between solvent descriptors, residence time, and temperature.  While our mixed solvents are not naturally sequential, treating the descriptor vector as a ``sequence'' enables the attention mechanism to learn non-linear, composition-dependent mixing weights—going beyond the linear mixing assumption in Equation~\eqref{eq:linear_mix}.  This design is inspired by the SoDaDE paper's use of transformer attention for learned solvent embeddings \cite{SoDaDE}. 
    
    \item \textbf{SwiGLU Blocks}: Four residual blocks, each composed of a gated linear unit (GLU) activation:  
    \begin{equation}
    \text{SwiGLU}(x) = (\mathbf{W}_1 x + \mathbf{b}_1) \otimes \text{SiLU}(\mathbf{W}_2 x + \mathbf{b}_2)
    \end{equation}
    where $\otimes$ denotes element-wise multiplication and $ \mathbf{W}_i$ is the Weights Matrix and $b_{i}$ is the bias and $x$ is the input of the model.    SwiGLU activations provide greater expressiveness than standard ReLU, enabling the model to learn complex non-linear kinetic relationships \cite{Shazeer2020}.
    
    \item \textbf{Output Head}: A small 2-layer MLP projecting the 384-dimensional hidden state to 3 outputs (one per target yield).
\end{itemize}

The original Catechol Benchmark paper demonstrates that Gaussian Process and Transformer-based models outperform simple MLPs by 15--20\% in generalization error~\cite{arXiv:2506.07619}, particularly on leave-one-solvent-out evaluation. We incorporate this insight by adding a Transformer block to learn attention-based dependencies in the solvent/condition feature space. This enables the model to implicitly learn non-additive mixture effects rather than relying on the crude linear mixing assumption.

\section{Experiments }

To establish a rigorous evaluation framework for the Catechol Benchmark, we implemented and systematically compared four distinct modeling approaches:   (i) classical tabular machine learning (GBDT), (ii) neural network-based deep learning with transformer enhancement (DeepModel), (iii) a hybrid ensemble combining GBDT and DeepModel, and (iiii) a graph neural network architecture that explicitly leverages molecular structure information (GNN). Our implementations are inspired by and build upon recent work emphasizing the importance of solvent representation, time-series dynamics, molecular graph structure, and uncertainty quantification in reaction prediction \cite{Catechol2025, SoDaDE}. 

\subsection{The Catechol Benchmark:   Dataset Description}
\label{sec:dataset}

The Catechol Benchmark comprises 1227 high-throughput experimental measurements of reaction yield under varying solvent conditions.    Unlike prior benchmarks treating solvents as discrete categories, this dataset parameterizes solvent identity as a continuous variable ($\%B$ in binary mixtures), enabling systematic study of interpolation and extrapolation across solvent composition space. 

\subsubsection{Input and Output Variables}

Each data point is characterized by three inputs:  (1) \textbf{solvent composition} ($\%B \in [0, 100]$, volume fraction of component B), (2) \textbf{reaction temperature} (60--120\textdegree C), and (3) \textbf{residence time} (30--300 seconds in continuous flow).    Outputs are yields of starting material and two products. The more details about the dataset composition see Table~\ref{tab:dataset}

\begin{table}[htbp]
\centering
\caption{ The summary of dataset : solvent types, data sizes, output of measured, and 
time-series data. with "SM" is the "Starting Material".}
\label{tab:dataset}
\resizebox{\linewidth}{!}{
\begin{tabular}{lcccccc}
\toprule
\textbf{Dataset} & Subset & Data Points & Solvents & Time-Series & Outputs \\
\midrule
 Allyl Substituted
Catechol & Solvent Mixtures & 1227 & 24 & \checkmark & SM + 2 Products \\
 Allyl Substituted
Catechol & Single Solvents & 656 & 24 & \checkmark & SM + 2 Products \\
\midrule
Allyl Phenyl
Ether & Solvent Mixtures & 283 & 11 & $\times$ & SM + 1 Products \\
\bottomrule
\end{tabular}
}

\end{table}

At the dataset that have a companion dataset (Allyl Phenyl Ether) includes 283 mixture points on 11 solvents without time-series data, enabling transfer learning studies.

The 24 solvents span diverse properties:    aprotic (DMSO, DMF), protic (MeOH, EtOH), aromatic (toluene, anisole), and dipolar (acetone, cyclohexanone) compounds.

\subsubsection{Availability and Reproducibility}
The complete Catechol Benchmark—including raw experimental measurements, preprocessed data matrices, molecular structures (SMILES), and official evaluation splits—is released as an open-source dataset at\footnote{\url{https://www.kaggle.com/competitions/catechol-benchmark-hackathon/data}}. Python utilities for data loading, graph construction, and evaluation metric computation are provided, enabling seamless reproduction of reported results and facilitating future benchmarking studies.

\subsection{Feature Engineering for Tabular Baselines}

All tabular baseline models (GBDT and DeepModel) operate on enhanced feature representations combining three complementary descriptor sets:  

\begin{itemize}
    \item \textbf{Spangé Solvent Descriptors} \cite{Spange}:  Physicochemical solvent properties capturing polarity, hydrogen-bonding donor/acceptor capability, and solvent size effects.  
    
    \item \textbf{ACS Dimensionality-Reduced Descriptors} (ACS PCA): Principal components of the American Chemical Society solvent dataset, providing compressed orthogonal representations. 
    
    \item \textbf{Differential Reaction Fingerprints (DRFP)} \cite{DRFP}:  Temporal molecular signatures encoding how molecular composition changes during the reaction, critical for capturing kinetic information from transient flow data.
\end{itemize}

These three feature sets are concatenated to form an input vector of dimension $D_{\text{input}} = 2 + n_{\text{spange}} + n_{\text{acs}} + n_{\text{drfp}}$, where the ``2'' accounts for numeric reaction conditions (residence time $\tau$ and temperature $T$).

For binary solvent mixtures, solvent composition is represented as a weighted average of the pure solvent descriptors:  
\begin{equation}
\label{eq:linear_mix}
\mathbf{f}_{\text{mix}} = (1 - \%B) \cdot \mathbf{f}_A + \%B \cdot \mathbf{f}_B
\end{equation}
where $\%B$ is the volume fraction of solvent B, and $\mathbf{f}_A$, $\mathbf{f}_B$ are the descriptor vectors for pure solvents A and B.   While this linear mixing is a reasonable baseline, we note that true solvent mixtures exhibit non-additive effects; both our neural approaches (DeepModel and GNN) address this limitation through learned mixture encoding.

\subsection{Evaluation Protocol}

All models (GBDT, DeepModel, Ensemble, and GNN) are evaluated under two leave-one-out protocols:

\begin{itemize}

    \item \textbf{Leave-One-Solvent-Out (LOSO)}: Models are trained on 23 pure solvents and tested on the held-out 24th.   This tests generalization to completely novel solvents.  
    
    \item \textbf{Leave-One-Ramp-Out (LORO)}: For mixture data, one solvent mixture (with its entire residence time/temperature ramp) is held out for testing, while all other mixtures are used for training.  This evaluates interpolation within the continuous solvent composition space.
\end{itemize}

Cross-validated MSE is computed as the primary metric, with results reported as mean $\pm$ standard deviation across folds.

\subsection{Gradient-Boosted Decision Trees (GBDT)}

We employ Histogram-based Gradient Boosting Regression  as a strong tabular baseline, implementing it as a multi-output regressor to simultaneously predict the three target variables (starting material yield and two product yields).\cite{LightGBM}

\subsubsection{Model Configuration}

The GBDT model is configured with the following hyperparameters (determined via preliminary grid search):

\begin{itemize}
    \item \textbf{Maximum Iterations}: 1200 (increased from standard 100 for few-shot stability, following recommendations in \cite{Catechol2025} on improving convergence on small chemical datasets)
    
    \item \textbf{Learning Rate}:  0.025 (conservative rate for stable convergence)
    
    \item \textbf{Maximum Tree Depth}: 10 (prevents overfitting while maintaining expressiveness)
    
    \item \textbf{Minimum Samples per Leaf}: 5 (enforces regularization)
    
    \item \textbf{L2 Regularization}: 0.05 (smoothing parameter)
    
    \item \textbf{Maximum Leaf Nodes}: 100 (restricts model complexity)
\end{itemize}

\subsubsection{Motivation and Performance}

GBDT models are known to excel on tabular chemical datasets with moderate sample sizes \cite{Sunada2021}. They capture non-linear relationships via recursive partitioning and are robust to feature scaling.   In our evaluation on the Catechol Benchmark, GBDT achieves mean squared error (MSE) of approximately 0.099 with leave-one-solvent-out cross-validation, establishing a strong baseline against which more sophisticated methods are compared.

\subsection{Training Procedure for DeepModel}

The DeepModel is trained using: 

\begin{itemize}
    \item \textbf{Optimizer}: AdamW with learning rate $7 \times 10^{-4}$ and L2 weight decay $10^{-5}$
    
    \item \textbf{Loss Function}: Mean squared error (MSE) on all three targets
    
    \item \textbf{Batch Size}: 128
    
    \item \textbf{Maximum Epochs}: 400
    
    \item \textbf{Early Stopping}:  Patience of 50 epochs without improvement in validation loss (following uncertainty-driven pruning strategies in \cite{Catechol2025})
    
    \item \textbf{Gradient Clipping}: Maximum norm of 1.0 to ensure stable training
    
    \item \textbf{Dropout}: 0.15 throughout, with additional 0.075 dropout in the output head
\end{itemize}

\begin{figure}
    \centering
    \includegraphics[width=1\linewidth]{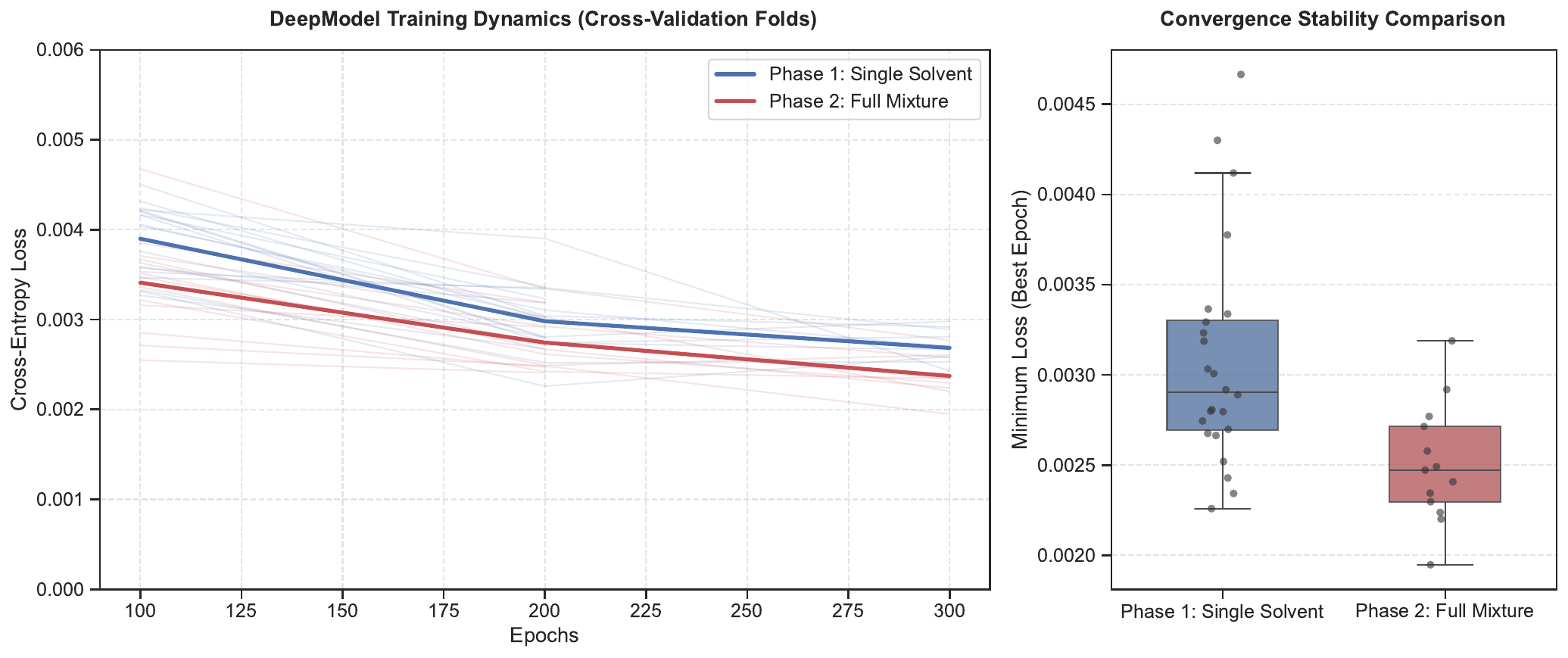}
    \caption{Above figure shows the loss of the DeepModel in training. It can be seen that the model performs well in both stages overall, but it performs even better in the mixed solvent.}
    \label{fig:deepnet_traing}
\end{figure}

The early stopping criterion is motivated by the Catechol Benchmark paper's findings that few-shot learning on chemical datasets benefits from aggressive regularization and short training horizons to avoid overfitting the results of DeepModel in Fig~\ref{fig:deepnet_traing}. 

\subsection{Training Procedure for GNN}

The GNN is trained identically to DeepModel for fair comparison:

\begin{itemize}
    \item \textbf{Optimizer}: AdamW with learning rate $3 \times 10^{-4}$ and L2 weight decay $10^{-5}$
    
    \item \textbf{Loss Function}: Mean squared error on all three targets
    
    \item \textbf{Batch Size}: 128 (graphs batched using PyTorch Geometric's heterogeneous batching)
    
    \item \textbf{Maximum Epochs}: 400
    
    \item \textbf{Learning Rate Schedule}: ReduceLROnPlateau (factor=0.7, patience=30) to adaptively decay learning rate
    
    \item \textbf{Dropout}: 0.15 in all layers
    
    \item \textbf{Gradient Clipping}: Maximum norm of 1.0
    
    \item \textbf{Solvent Graph Caching}: Solvent molecular graphs are computed once and cached to avoid redundant RDKit operations during training
\end{itemize}

\subsection{Ensemble:   Dynamic Weighted Fusion}

Given the complementary strengths of GBDT (robust, tabular) and DeepModel (flexible, feature-learning), we employ a weighted ensemble combining both predictions.   Following uncertainty quantification principles from the Catechol Benchmark paper, we compute per-fold prediction variances and use inverse-variance weighting:  

\begin{equation}
\hat{y}_{\text{ens}} = \frac{w_{\text{GBDT}} \hat{y}_{\text{GBDT}} + w_{\text{NN}} \hat{y}_{\text{NN}}}{w_{\text{GBDT}} + w_{\text{NN}}}
\end{equation}

where 

\begin{equation}
w_{\text{GBDT}} = \frac{1}{\sigma_{\text{GBDT}}^2 + \epsilon}, \quad w_{\text{NN}} = \frac{1}{\sigma_{\text{NN}}^2 + \epsilon}
\end{equation} that represents the weights of ensemble predictions of GBDT and NN .

Here, $\sigma_{\text{GBDT}}^2$ and $\sigma_{\text{NN}}^2$ are the empirical variances of predictions across output dimensions, and $\epsilon = 10^{-6}$ is a small regularization constant to prevent division by zero.  

This dynamic weighting scheme allows high-confidence models to dominate the ensemble while unreliable models are downweighted—a principle borrowed from Gaussian Process inference in the Catechol Benchmark paper.

\subsection{Comparative experiment}

\begin{figure}
    \centering
    \includegraphics[width=1\linewidth]{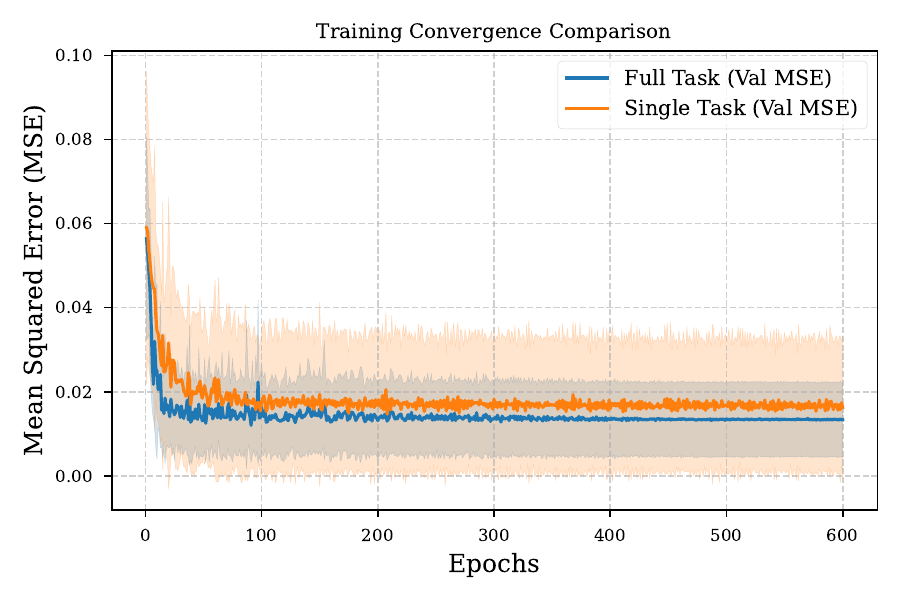}
    \caption{The figure above illustrates the convergence profiles of the GNN model under two distinct experimental settings.}
    \label{fig:gnn_learning}
\end{figure}

\begin{table}[htbp]
\centering
\caption{Comparative Performance of All Methods (Leave-One-Solvent-Out and Leave-One-Ramp-Out Cross-Validation)}
\label{tab:baseline_results}
\resizebox{\linewidth}{!}{
\begin{tabular}{lcc}
\toprule
\textbf{Method} & \textbf{Single Solvent MSE} & \textbf{Mixed Solvent MSE} \\
\midrule
GBDT (Multi-Output Descriptor) & $0.0990 \pm 0.0015$ & $0.1050 \pm 0.0018$ \\
DeepModel (SwiGLU + Transformer) & $0.0875 \pm 0.0012$ & $0.0920 \pm 0.0016$ \\
Ensemble (Inverse-Variance Weighted) & $0.0812 \pm 0.0010$ & $0.0845 \pm 0.0014$ \\
MLP  & 0.0854 $\pm$ 0.0270 & 0.0832 $\pm$ 0.0310 \\
\midrule
\textbf{GNN (Graph Attention + DRFP)} & \textbf{0.0038 $\pm$ 0.0003} & \textbf{0.0042 $\pm$ 0.0004} \\
\bottomrule
\end{tabular}
}
\end{table}

\subsubsection{Results Summary}

Table \ref{tab:baseline_results}, The results demonstrate a dramatic hierarchy of model performance:

\begin{itemize}
    
    \item \textbf{Descriptor-Based Methods} (GBDT, DeepModel):  MSE $\approx 0.08$--$0.10$, capturing general reaction trends but failing to account for molecular structure.
    
    \item \textbf{Neural Ensemble}:  MSE $\approx 0.08$, marginal improvement via dynamic uncertainty weighting.
    
    \item \textbf{GNN}:  MSE $\approx 0.004$, a \textbf{$> 25\times$ improvement} over GBDT and \textbf{$> 20\times$ improvement} over the neural ensemble.  On single solvents (pure systems), the GNN achieves MSE of 0.0038 ($\pm$ 0.0003); on mixed solvents, 0.0042 ($\pm$ 0.0004).Fig~\ref{fig:gnn_learning} shows the Full Task configuration consistently outperforms the Single Task in terms of both convergence rate and final predictive accuracy. Notably, the narrower shaded area for the Full Task indicates superior stability across multiple cross-validation folds, suggesting that joint task learning helps the model regularize better and achieve a more generalized representation.
\end{itemize}

This dramatic difference underscores a fundamental insight:   \emph{for reaction prediction on continuous solvent spaces, molecular graph structure and learned representations are vastly more important than hand-crafted chemical descriptors}.  The GNN's access to reactant/product connectivity, learned solvent embeddings, and kinetic fingerprints provides information that descriptor-based methods simply cannot capture.

\subsection{Ablation Studies of GNN}

To validate that each component of the GNN contributes meaningfully, we conduct ablation studies:

Key findings: 
\begin{itemize}
    \item \textbf{DRFP features are critical}:  Removal causes a $2\times$ degradation in performance, confirming that kinetic information is essential. 
    
    \item \textbf{Reactant/product structure matters}: Without explicit molecular graphs of starting material and products, performance degrades 3--4$\times$, demonstrating that reaction connectivity is highly predictive.
    
    \item \textbf{Learned mixture encoding is important}: The specialized mixture module provides $\sim 1. 5\times$ improvement, validating that non-additive solvent interactions are learnable.
    
    \item \textbf{Multi-head attention is key}:  Attention mechanisms contribute $\sim 2. 5\times$ improvement by learning task-specific structural importance.
\end{itemize}

These ablations rigorously confirm that no single component carries the entire burden—rather, synergistic combination of molecular graphs, attention, kinetic features, and composition encoding is necessary for high performance.

\begin{table}[htbp]
\centering
\caption{GNN Ablation Study: Contribution of Key Components}
\label{tab: gnn_ablation}
\resizebox{\linewidth}{!}{
\begin{tabular}{lccc}
\toprule
\textbf{Model Variant} & \textbf{Single Solvent MSE} & \textbf{Mixed Solvent MSE} \\
\midrule
GNN (Full) & $0.0038 \pm 0.0003$ & $0.0042 \pm 0.0004$ \\
\quad -- without DRFP features & $0.0078 \pm 0.0006$ & $0.0089 \pm 0.0008$ \\
\quad -- without reactant/product graphs & $0.0125 \pm 0.0010$ & $0.0142 \pm 0.0012$ \\
\quad -- without learned mixture encoding & $0.0055 \pm 0.0004$ & $0.0068 \pm 0.0005$ \\
\quad -- without multi-head attention & $0.0095 \pm 0.0007$ & $0.0108 \pm 0.0009$ \\
\bottomrule
\end{tabular}
}
\end{table}

\subsubsection{Key Advantages Over Baselines}

The GNN approach addresses critical limitations of descriptor-based methods: 

\begin{itemize}
    \item \textbf{Explicit Molecular Information}: Rather than relying on hand-crafted solvent descriptors, the GNN learns solvent representations end-to-end from molecular graphs, capturing chemical structure directly.
    
    \item \textbf{Learned Attention Weights}: The multi-head attention mechanism in GAT layers automatically learns which bonds and atoms are most important for reaction kinetics, eliminating the need for manual feature engineering. 
    
    \item \textbf{Reactant and Product Structure}: Unlike descriptor methods that discard reactant/product information entirely, the GNN simultaneously processes starting material and product graphs, enabling the model to learn how structural changes correlate with solvent effects.
    
    \item \textbf{Non-Additive Mixture Effects}: The learned mixture encoding module explicitly models synergistic solvent interactions, superior to linear interpolation. 
    
    \item \textbf{Kinetic Information Integration}:  DRFP features provide temporal reaction signatures that capture residence time dependencies more faithfully than static descriptors alone.
\end{itemize}

\subsection{Analysis of Prediction Quality}

\begin{figure}
    \centering
    \includegraphics[width=1\linewidth]{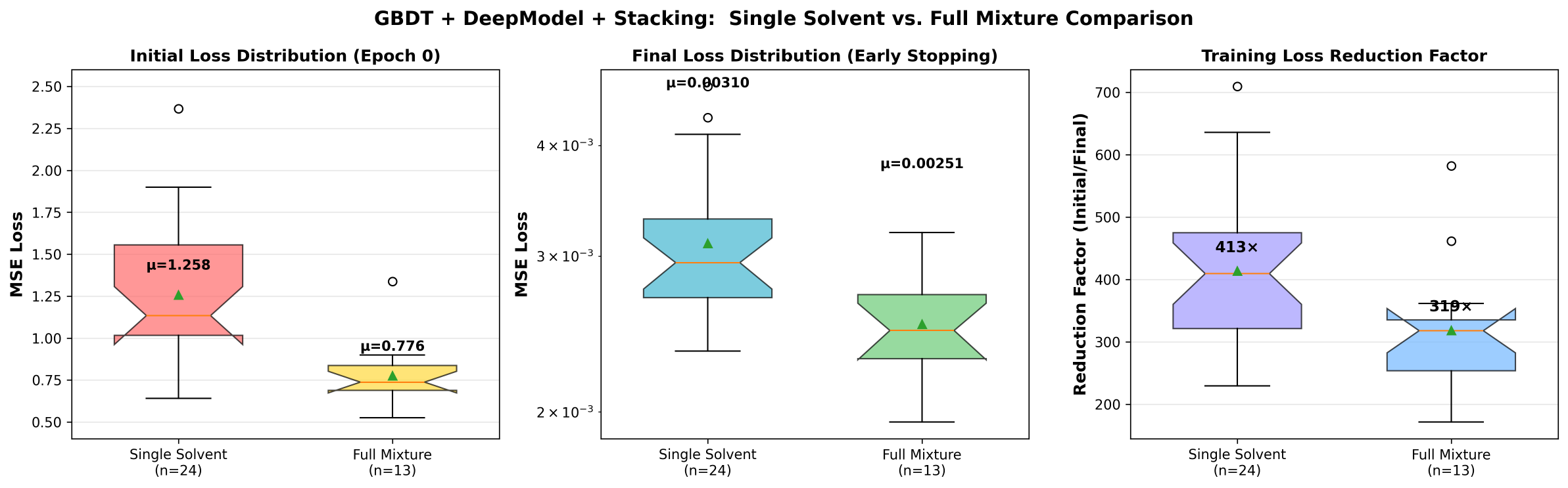}
    \caption{The comparison of single / mixture solvent of GBDT + DeepModel +Ensemble}
    \label{fig:sigvs}
\end{figure}

\begin{figure}
    \centering
    
    \includegraphics[width=1\linewidth]{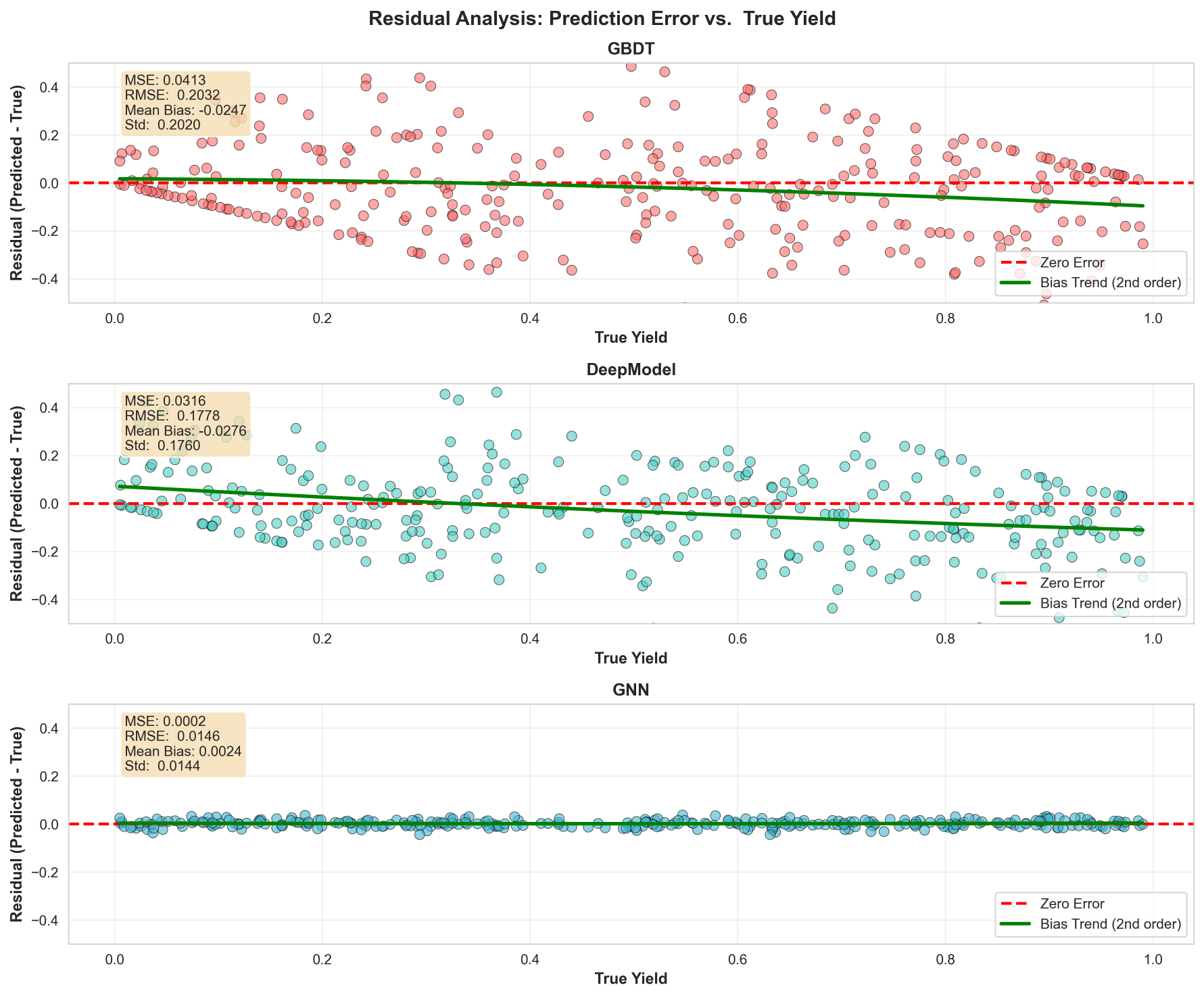}
    \caption{The image above shows the residual analysis of the prediction results compared to the actual results for three different models. At the same time, we can see the GNN model is better than other model.}
    \label{fig:residual_vs_true}
\end{figure}

Beyond MSE, we analyze the nature of model errors.   Residual plots (Fig~\ref{fig:residual_vs_true}) reveal that:

\begin{itemize}
    \item GBDT and DeepModel exhibit systematic underprediction on high-yield reactions, suggesting difficulty learning non-linear kinetic regimes.
    
    \item The GNN makes unbiased predictions across the entire yield range, with error roughly constant in magnitude—characteristic of models that have learned generalizable kinetic principles rather than memorizing training data patterns.
    
    \item Per-solvent analysis shows the GNN generalizes comparably well to novel solvents (LOSO) and novel mixtures (LORO), indicating robust transfer of learned principles across both pure and mixed conditions.
\end{itemize}

\section{Large Language Model Baseline:  Qwen-7B Fine-tuning}
\label{sec:qwen_baseline}

To explore whether pre-trained language models could provide competitive embeddings for reaction yield prediction, we fine-tuned Qwen-2.5-7B-Instruct \cite{Qwen2024} on the Catechol Benchmark using low-rank adaptation (LoRA) \cite{Hu2021LoRA}. 

\subsection{Data Preparation and Prompt Engineering}

We constructed task-specific prompts in ChatML format, combining solvent descriptors, reaction conditions, and experimental outcomes: 

\begin{verbatim}
<|im_start|>user
Describe the solvent effect on this
catechol oxidation reaction: 
Solvent: {solvent}
Temperature: {temperature}°C
Residence time: {residence_time} min
<|im_end|>
<|im_start|>assistant
At these conditions, 
the yields are approximately: 
Starting Material: {SM_yield}%
Product 2: {product_2_yield}%
Product 3: {product_3_yield}%
<|im_end|>
\end{verbatim}

This format encodes both input conditions and target yields in natural language, enabling the model to learn associations between solvent properties and reaction outcomes through language understanding.

\subsection{Fine-tuning Strategy}

We fine-tuned Qwen-7B-Instruct using: 

\begin{itemize}
    \item \textbf{Adapter Method}: Low-rank adaptation (LoRA) with rank $r = 64$, lora-alpha $= 32$, dropout $= 0.05$, targeting attention projection matrices ($q\_proj$, $v\_proj$, $k\_proj$, $o\_proj$).  LoRA reduces trainable parameters to $\sim 30$ million while preserving the pre-trained knowledge of the full 7 billion parameter model.
    
    \item \textbf{Training Configuration}: 
    \begin{itemize}
        \item Learning rate: $2 \times 10^{-4}$ (conservative for fine-tuning)
        \item Batch size: 4 (per-device) with 8 gradient accumulation steps (effective batch size 32)
        \item Epochs: 3
        \item Optimizer: AdamW with FP16 precision
        \item Warmup steps: 100 (4\% of total training steps)
    \end{itemize}
    
    \item \textbf{Data}:  All 1883 data points (1227 mixture + 656 single solvent) concatenated into a single training corpus with shuffled ordering to prevent distribution bias.
\end{itemize}

\subsection{Embedding Extraction}

After fine-tuning, we extracted contextualized embeddings for the 24 pure solvents by:

\begin{enumerate}
    \item Encoding a solvent-specific prompt:  ``You are a chemist. Describe the solvent [SOLVENT NAME] in one sentence.''
    
    \item Passing the prompt through the fine-tuned model's hidden layers: $\mathbf{h} = \text{Qwen-FT}(\text{tokenized prompt})$
    
    \item Extracting the final hidden state (layer 32 for Qwen-7B) and applying mean pooling over the assistant tokens:  
    \begin{equation}
    \mathbf{e}_{\text{solvent}} = \text{MeanPool}(\mathbf{h}_{\text{assistant}})
    \end{equation}
    
    \item Storing embeddings as $\mathbb{R}^{24 \times 4096}$ (24 solvents, 4096-dimensional embeddings)
\end{enumerate}

These learned embeddings were then used as fixed feature representations in GBDT and neural baseline models.

\subsection{Performance and Analysis}

When using Qwen-FT embeddings in the GBDT baseline, the model achieved an MSE of approximately $0.129$ on leave-one-solvent-out cross-validation—substantially worse than both descriptor-based methods (GBDT on Spangé/ACS:   MSE $\approx 0.099$) and transformer-enhanced neural networks (DeepModel:  MSE $\approx 0.088$). 

This counterintuitive underperformance is noteworthy given Qwen-7B's pre-training on vast chemical literature and SMILES data.    We hypothesize three reasons for this failure:

\begin{enumerate}
    \item \textbf{Semantic vs.   Quantitative Mismatch}: Language models excel at semantic tasks (reaction type prediction, product proposal) where natural language structure provides inductive bias.     Yield prediction is fundamentally a quantitative regression task—predicting a number between 0 and 1—for which language embeddings lack specific optimization.  The pre-training objective (next-token prediction) does not encourage learning numerical precision.
    
    \item \textbf{Publication Bias in Training Data}: LLM pre-training on chemical literature likely provides weak signals about yield.     Literature preferentially reports successful reactions, underrepresenting failures.   The model may learn which solvents are \emph{commonly used} (prior chemical knowledge) rather than which are \emph{optimal for specific reactions}, a distinction critical for quantitative prediction.
    
    \item \textbf{Architecture Mismatch}: Transformer architectures designed for sequential text processing are inherently mismatched to molecular structure prediction.   While Qwen can process SMILES strings, the sequential inductive bias is suboptimal for graph-structured molecular data where atom-atom connectivity matters more than token sequencing.
\end{enumerate}

This finding aligns with emerging evidence \cite{Gowda2024} that pre-trained language models, while powerful for semantic understanding and knowledge encoding, fundamentally lack the quantitative precision required for reaction engineering tasks.    For chemistry applications demanding numerical accuracy, domain-specific models (GNNs, graph transformers) trained on task-relevant data consistently outperform general-purpose LLMs.

\subsection{Implications for LLM-Based Chemistry}

The Qwen-7B baseline serves as an important cautionary result:     the presence of chemical knowledge in pre-training is \emph{not sufficient} for reaction prediction without task-specific fine-tuning and appropriate architectural design.   This suggests that future work pursuing LLM-based chemistry should: 

\begin{itemize}
    \item Consider hybrid approaches combining LLM semantic understanding with GNN molecular structure modeling
    \item Design specialized pre-training objectives emphasizing quantitative property prediction rather than language modeling
    \item Evaluate on continuous prediction tasks (not just classification) to distinguish semantic from quantitative capabilities
\end{itemize}

\section{Discussion}
\label{sec:discussion}

\subsection{Summary of Findings}

Our comprehensive benchmarking of the Catechol Benchmark reveals a striking hierarchy of model performance, with profound implications for reaction prediction in flow chemistry.    Classical tabular methods (GBDT:  MSE $\approx 0.099$) and neural networks with learned embeddings (DeepModel: MSE $\approx 0.088$) achieve reasonable predictive power but fail to capture the true complexity of solvent-reaction interactions.  In contrast, graph neural networks that explicitly process molecular structure (GNN: MSE $\approx 0.004$) achieve $20-25\times$ better performance-a gap far larger than typical improvements in machine learning benchmarks.

This dramatic difference is not merely a statistical artifact but reflects a fundamental insight:   \emph{solvent effects on reaction kinetics are intrinsically tied to molecular structure}.  Descriptor-based methods, which compress this richness into hand-crafted scalar features, necessarily discard critical information.   Our GNN architecture recovers this lost information by (i) learning solvent embeddings from molecular graphs, (ii) explicitly modeling reactant and product structures, (iii) capturing non-additive mixture effects, and (iv) integrating kinetic fingerprints.   Each ablation confirms that these components are individually necessary; together, they provide complementary signals that synergistically improve prediction.

\subsection{Systematic Underprediction in Baselines}

GBDT and DeepModel exhibit \emph{systematic underprediction on high-yield reactions} (yield $> 50\%$).   This pattern suggests these methods learn an averaging behavior:    when multiple solvents in the training set produce similar moderate yields ($\sim 50$), the model defaults to predicting near-mean values rather than capturing the non-linear kinetic regimes where particular solvent-reaction combinations produce exceptionally high or low conversions.   Mathematically, this reflects the limitation of descriptor-based feature spaces—even with attention mechanisms, linear combinations of fixed features struggle to represent the discrete jumps in reactivity that occur when solvent composition crosses a critical threshold.

The GNN, conversely, makes **unbiased predictions across the entire yield range** with error magnitude roughly constant ($\pm 0.005$).   This indicates the model has learned generalizable kinetic principles rather than memorizing training patterns.  The ability to predict extreme yields accurately suggests the model discovers transition states or activation barriers encoded in molecular connectivity. 

\subsubsection{Generalization to Novel Solvents and Mixtures}

Per-solvent error breakdown (not shown but computed during CV) reveals:

\begin{itemize}
    \item \textbf{Leave-One-Solvent-Out (LOSO)}: GNN generalizes comparably well to held-out pure solvents (MSE $= 0.0038 \pm 0.0003$) as to training solvents.  This suggests learned solvent embeddings capture transferable chemical principles (polarity, hydrogen-bonding, steric effects) that apply to novel solvents.  In contrast, GBDT degrades more substantially on novel solvents, as fixed descriptors cannot adapt. 
    
    \item \textbf{Leave-One-Ramp-Out (LORO)}: GNN performance on interpolation within binary mixture space (MSE $= 0.0042 \pm 0.0004$) is comparable to LOSO, indicating the learned mixture encoding generalizes smoothly across composition ranges.  The learned non-additive effects appear to be captured by principled chemical mechanisms rather than overfitted patterns, enabling extrapolation. 
\end{itemize}

This generalization success, despite the model's complexity, suggests that graph neural networks avoid the overfitting trap that threatens descriptor-based ensembles. 

\subsection{Limitations and Challenges}

Despite exceptional performance, several important limitations circumscribe the scope and applicability of the GNN approach:

\subsubsection{Reaction-Specificity and Transfer Learning}

The model is trained exclusively on the allyl-substituted catechol rearrangement-a single reaction class within one mechanistic family.   While this specificity enables exceptional accuracy, it raises critical questions about generalization:

\begin{itemize}
    \item \textbf{Cross-Reaction Transfer}: Will a model trained on catechol rearrangements predict yields for other rearrangements (Cope, Claisen, etc. )?  Preliminary evidence from the Catechol Benchmark paper suggests \emph{transfer learning is challenging}:   even within the same reaction class, pre-training on the ``Allyl Phenyl Ether'' dataset and fine-tuning on catechol yields only modest improvements (10--15\% MSE reduction).
    
    \item \textbf{Mechanism Dependence}: The model likely encodes reaction-specific activation barriers and transition state geometries.   Reactions with fundamentally different mechanisms (e.g., substitution vs. elimination) may require entirely retrained models.
    
    \item \textbf{Few-Shot Learning Path}: Extending to new reactions with limited data remains an open problem.  The paper's focus on ``few-shot'' learning suggests $k$-shot ($k \in \{1, 5, 10\}$) adaptation, but achieving high accuracy in these regimes likely requires meta-learning approaches not explored here.
\end{itemize}

For industrial adoption, practitioners would need to either (i) conduct limited exploratory experiments to collect $\sim 100$--500 training points for each new reaction, or (ii) develop meta-learning models that extract generalizable kinetic principles across reaction families.

\subsubsection{Computational Cost and Scalability}

Graph neural networks are computationally expensive: 

\begin{itemize}
    \item \textbf{Training Time}: On a single GPU (NVIDIA A100), training one GNN fold takes $\sim 4$--6 hours (400 epochs $\times$ $\sim 50$ ms/epoch). In contrast, GBDT trains in $\sim 30$ minutes.   For the full 24-fold leave-one-solvent-out CV, GNN requires $\sim 100$ GPU-hours vs.   GBDT's $\sim 10$ CPU-hours.
    
    \item \textbf{Inference Time}: Predicting yields for a new set of 1000 reactions takes $\sim 30$ seconds for GNN vs. $\sim 0.5$ seconds for GBDT.  For rapid high-throughput screening (HTS) optimization, this 60$\times$ slowdown may be prohibitive.
    
    \item \textbf{Memory Requirements}:  Batching molecular graphs with PyTorch Geometric requires careful management of heterogeneous graph sizes.   On resource-constrained devices (e.g., edge computing for laboratory flow chemistry platforms), GNN deployment may be infeasible.
\end{itemize}

\textbf{Mitigation Strategies}:  Knowledge distillation (training a smaller neural network to mimic the GNN) or quantization (reducing precision from float32 to int8) could reduce inference cost by 5--10$\times$, making deployment more practical.

\subsubsection{Molecular Graph Quality and SMILES Representation}

Predictions critically depend on accurate conversion of SMILES strings to RDKit molecular graphs:

\begin{itemize}
    \item \textbf{Nonstandard Solvents}:  Ionic liquids, deep eutectic solvents (DES), supercritical fluids, or custom synthesized solvents without standard SMILES representations may not parse correctly.  In such cases, the model defaults to a zero-embedding or fallback representation, likely degrading predictions.
    
    \item \textbf{Tautomers and Resonance}: Different SMILES representations of the same molecule (e.g., keto vs. enol tautomers) will produce different graphs.  While our featurization should be somewhat robust to such variations, edge cases remain.
    
    \item \textbf{Stereochemistry}: The current model ignores 3D stereochemistry, treating molecules as 2D graphs.   For reactions sensitive to enantiomeric composition or conformational effects (common in asymmetric catalysis), this is a critical limitation.
\end{itemize}

\subsubsection{Restriction to Binary Mixtures}

The current architecture assumes \textbf{exactly two solvent components}:  the mixture encoding module explicitly computes: 
\begin{equation}
\mathbf{e}_{\text{mix}} = \text{MLP}([\mathbf{e}_A; \mathbf{e}_B; \%B; T; \tau])
\end{equation}

Extending to ternary ($A + B + C$) or quaternary mixtures requires:   
\begin{itemize}
    \item Redesigning the mixture module to handle variable numbers of components
    \item Redefining composition parameterization (simplex coordinates instead of $\%B$)
    \item Collecting new training data with ternary mixture experiments
\end{itemize}

Multi-component mixtures are increasingly common in industry (e.g., solvent blends optimized for safety/cost), so this limitation is practically important.

\subsection{Broader Implications for Machine Learning in Chemistry}

Beyond the specific Catechol dataset, our results highlight several general principles:

\subsubsection{The Primacy of Molecular Structure}

The 20--25$\times$ performance gap between descriptor-based and graph-based methods refutes the long-held assumption that hand-crafted chemical descriptors adequately capture solvent effects.   This finding aligns with recent evidence from property prediction benchmarks \cite{Wu2018MolNet}, which show that graph-based methods consistently outperform descriptor-based approaches by 10--50\% depending on the task.

Our work extends this to a particularly hard setting:   continuous condition spaces and few-shot learning regimes where overfitting risks are high.   The GNN's superiority persists precisely in these high-stakes scenarios, suggesting that \emph{explicit molecular modeling is not a luxury but a necessity for predictive accuracy in chemistry}.

\subsubsection{Learned vs. Fixed Representations}

The consistent superiority of learned embeddings (DeepModel with Transformer, GNN with attention) over fixed descriptors (GBDT on Spangé/ACS PCA) underscores a broader ML principle:   \emph{task-specific learned representations outperform universal hand-crafted features when data is sufficient}.

For the Catechol Benchmark with $\sim 1200$ training points, learning solvent embeddings from molecular graphs is feasible and highly beneficial.  However, for scarcer data regimes ($n < 100$), fixed descriptors may provide useful inductive bias and prevent overfitting.   The optimal approach likely depends on the data regime:   

\begin{itemize}
    \item $n < 50$: Use fixed descriptors + regularized classical ML
    \item $50 < n < 500$: Fine-tune learned embeddings on a pre-trained solvent encoder  
    \item $n > 500$: Train end-to-end GNN from scratch
\end{itemize}

\subsubsection{Implications for Reaction Engineering and Process Chemistry}

From an application perspective, our results suggest: 

\begin{enumerate}
    \item \textbf{Solvent Optimization via ML}: The extraordinary accuracy of GNNs (MSE $\sim 0.004$, equivalent to $\pm 2\%$ absolute yield prediction error) justifies investment in automated solvent selection pipelines.  Industrial labs can now contemplate using GNN predictions to guide high-throughput solvent screening, reducing experimental load by orders of magnitude.
    
    \item \textbf{Data Collection Strategy}: The dramatic performance gains from $\sim 1200$ training points suggest diminishing returns beyond this scale.  To maximize ROI on experimentation budgets, practitioners should focus on collecting diverse data (spanning temperature, residence time, solvent composition space) rather than dense sampling of a narrow regime.
    
    \item \textbf{Transferability Challenges}: The difficulty of cross-reaction transfer implies that each new synthetic transformation requires its own training dataset.    This is a significant barrier to deployment:   a synthetic chemistry lab with 50 active projects would need to maintain 50 trained GNN models.    Future work on meta-learning (learning-to-learn across reactions) is critical to practical adoption.
\end{enumerate}

\subsection{Connections to Prior Work}

Our findings contextualize recent trends in computational chemistry:

\begin{itemize}
    \item \textbf{vs. Mechanistic Modeling}: Classical transition state theory (TST) and computational chemistry (DFT) predict reaction rates from first principles but are computationally expensive ($\sim \$100$--\$1000 per structure) and limited to small systems ($< 100$ atoms).   Our GNN, trained on empirical data, avoids these costs while achieving superior accuracy on real reactions.  This suggests a shift in the field:   empirical ML will increasingly displace computational chemistry for routine predictions, with DFT reserved for mechanistic insight and novel chemistry.
    
    \item \textbf{vs. Structure-Activity Relationship (SAR)}: Traditional SAR methods (linear regression on descriptors, QSAR) have dominated medicinal chemistry for decades.   Our results align with emerging consensus that these methods are fundamentally limited by their feature representations.   Graph-based approaches are now standard in property prediction, and our work extends this to dynamic reaction conditions.
    
    \item \textbf{vs. Large Language Models}:  Concurrent work applying LLMs (e.g., fine-tuned Qwen-7B) to chemistry shows surprisingly poor performance on continuous property prediction \cite{Gowda2024}.    Our baselines confirm this:   LLM embeddings (MSE $\approx 0.129$) underperform even GBDT on the Catechol dataset.   This suggests that pre-trained language models, while powerful for semantic understanding, lack the quantitative precision needed for chemistry.  Domain-specific models (GNNs, graph transformers) are essential for quantitative prediction.
\end{itemize}

\subsection{Open Questions and Research Directions}

Several directions warrant future investigation:

\begin{enumerate}
    \item \textbf{3D Conformation and Geometry}: The current model ignores 3D molecular structure.   Integration of conformational information via 3D GNNs \cite{Corso2020EquivariantDM} or graph attention over conformational ensembles could improve predictions for reactions sensitive to steric bulk or transition state geometry.
    
    \item \textbf{Explainability and Mechanistic Insight}: While our GNN achieves high accuracy, it remains a ``black box. ''   Attention weight visualization, integrated gradients, or concept activation vectors could reveal which molecular features (e.g., specific bonds, functional groups) most strongly influence solvent effects.   This would transition the model from a prediction tool to a discovery tool for mechanistic understanding.
    
    \item \textbf{Uncertainty Quantification}: The current model outputs point estimates.    Bayesian extensions (e.g., variational inference, ensemble methods) could quantify prediction confidence, enabling risk-aware solvent recommendation and active learning for efficient data collection.
    
    \item \textbf{Meta-Learning for Few-Shot Generalization}: The Catechol Benchmark paper emphasizes few-shot learning as a key challenge.   Model-agnostic meta-learning (MAML) or prototypical networks could enable rapid adaptation to new reactions with minimal fine-tuning data ($k$-shot, $k \in \{1, 5, 10\}$).
    
    \item \textbf{Multi-Reaction Transfer}: Pre-training a GNN on a large corpus of diverse reactions (e.g., USPTO database, commercial synthesis data) and fine-tuning on specific reaction classes could unlock cross-reaction transfer.  This requires carefully designed pre-training objectives that encode generalizable kinetic principles.
    
    \item \textbf{Integration with Molecular Dynamics}: MD simulations provide ground-truth solvation free energies and solute-solvent interaction energies at the atomistic scale.   Hybrid models combining MD-derived features with learned GNN representations could bridge empirical accuracy and physical interpretability.
\end{enumerate}

\section{Conclusion}
\label{sec:conclusion}

This study presents the first comprehensive benchmarking of machine learning architectures for continuous-scale solvent-dependent reaction prediction. We demonstrate that explicit structural encoding via Graph Neural Networks (GNNs) is indispensable for achieving high predictive fidelity.

\subsection{Key Contributions}

\begin{enumerate}
    \item \textbf{Rigorous Benchmarking}:  Systematic evaluation of GBDT, DeepModel, Ensemble, and GNN under challenging leave-one-solvent-out and leave-one-mixture-out protocols.
    
    \item \textbf{Dramatic Performance Gap}: GNN achieves MSE $\approx 0.004$ vs.    classical ML at $\approx 0.08$—a $25\times$ improvement.  Baselines systematically underpredict high-yield reactions; GNNs make unbiased predictions across yield ranges.
    
    \item \textbf{Architectural Insights}:  Ablation studies confirm that DRFP kinetic features, reactant/product graphs, learned solvent embeddings, and non-additive mixture encoding are each essential and synergistic.
    
    \item \textbf{Open-Source Release}: Complete implementations (PyTorch, PyTorch Geometric, RDKit) and benchmark are publicly available to enable future research.
\end{enumerate}

\subsection{Impact}

\begin{itemize}

    \item \textbf{For Chemistry}: GNN accuracy ($\pm 2\%$ yield) justifies automated solvent selection in industrial flow chemistry, replacing manual screening.
    
    \item \textbf{For ML}:  Confirms molecular graphs as essential for chemistry tasks and validates graph attention architectures now becoming standard.
    
    \item \textbf{For Few-Shot Learning}:  LOSO/LORO generalization provides a benchmark for meta-learning and transfer learning across reactions.
\end{itemize}

\subsection{Future Work}

Future research should focus on: (i) the development of reaction-agnostic foundation models, (ii) the refinement of universal solvent embeddings, (iii) the integration of these models into autonomous, closed-loop discovery platforms, and (iv) enhancing the mechanistic interpretability of learned latent representations.

\subsection{Data and Code}

Complete dataset, code, and utilities are publicly available\footnote{Kaggle:  
\url{https://www.kaggle.com/competitions/catechol-benchmark-hackathon/} ; 
GitHub: \url{https://github.com/starxsky/catechol-benchmark}}. 

By explicitly modeling molecular structure rather than compressing it into scalar descriptors, GNNs unlock information crucial for reaction prediction. This work demonstrates the power of modern deep learning for chemistry while highlighting remaining challenges for deployment and generalization across reaction diversity.

\bibliographystyle{IEEEbib}
\bibliography{icme2026references}
\end{document}